\documentclass[letterpaper]{article} 
\usepackage{aaai25}  
\usepackage{times}  
\usepackage{helvet}  
\usepackage{courier}  
\usepackage[hyphens]{url}  
\usepackage{graphicx} 
\usepackage{natbib}  
\usepackage{caption} 
\usepackage{algorithm}
\usepackage{algorithmic}
\usepackage{subcaption} 
\usepackage{newfloat}
\usepackage{listings}
\DeclareCaptionStyle{ruled}{labelfont=normalfont,labelsep=colon,strut=off} 
\floatstyle{ruled}
\newfloat{listing}{tb}{lst}{}
\floatname{listing}{Listing}
\usepackage{amssymb}
\usepackage{booktabs}
\usepackage{xcolor}
\usepackage{multirow}
\usepackage{arydshln}
\usepackage{bm}
\usepackage{amsmath}
\usepackage[capitalize]{cleveref}
\newcommand{\ie}{\textit{i}.\textit{e}.}
\newcommand{\eg}{\textit{e}.\textit{g}.}
\newcommand{\etal}{\textit{et al}.}
\title{CLIP-PCQA: Exploring Subjective-Aligned Vision-Language Modeling \\for Point Cloud Quality Assessment}
\author{
    Yating Liu\equalcontrib,
    Yujie Zhang\equalcontrib,
    Ziyu Shan,
    Yiling Xu\thanks{Corresponding author.}
}

\affiliations{
    Cooperative Medianet Innovation Center, Shanghai Jiao Tong University, Shanghai, China\\
    \{Olivialyt, yujie19981026, shanziyu, yl.xu\}@sjtu.edu.cn
}

\begin{document}
\maketitle

\begin{abstract}
In recent years, No-Reference Point Cloud Quality Assessment (NR-PCQA) research has achieved significant progress. However, existing methods mostly seek a direct mapping function from visual data to the Mean Opinion Score (MOS), which is contradictory to the mechanism of practical subjective evaluation. To address this, we propose a novel language-driven PCQA method named CLIP-PCQA. Considering that human beings prefer to describe visual quality using discrete quality descriptions (\eg, ``excellent" and ``poor") rather than specific scores, we adopt a retrieval-based mapping strategy to simulate the process of subjective assessment. More specifically, based on the philosophy of CLIP, we calculate the cosine similarity between the visual features and multiple textual features corresponding to different quality descriptions, in which process an effective contrastive loss and learnable prompts are introduced to enhance the feature extraction. Meanwhile, given the personal limitations and bias in subjective experiments, we further covert the feature similarities into probabilities and consider the Opinion Score Distribution (OSD) rather than a single MOS as the final target. Experimental results show that our CLIP-PCQA outperforms other State-Of-The-Art (SOTA) approaches.

\begin{links}
    \link{Code}{https://github.com/Olivialyt/CLIP-PCQA}
\end{links}

\end{abstract}
\section{Introduction}
\label{sec:intro}
Point clouds are essentially collections of points scattered in 3D space, where each point has the corresponding geometry coordinates and other attributes. With its widespread applications in various fields (\eg, autonomous driving~\cite{cui2021deep} and virtual/augmented reality~\cite{park2008multiple}), point clouds have become one of the most typical representations of 3D objects or scenes. However, many visual artifacts could be introduced into point clouds during processing stages, such as capturing, compression, transmission, and rendering. In order to ensure the Quality Of Experience (QoE), it is necessary to develop effective metrics that accurately measure the perceptual distortion, especially in the common No-Reference (NR) situation where ``perfect" reference point clouds are not available.

\begin{figure}
    \centering
    \includegraphics[width = 1\linewidth]{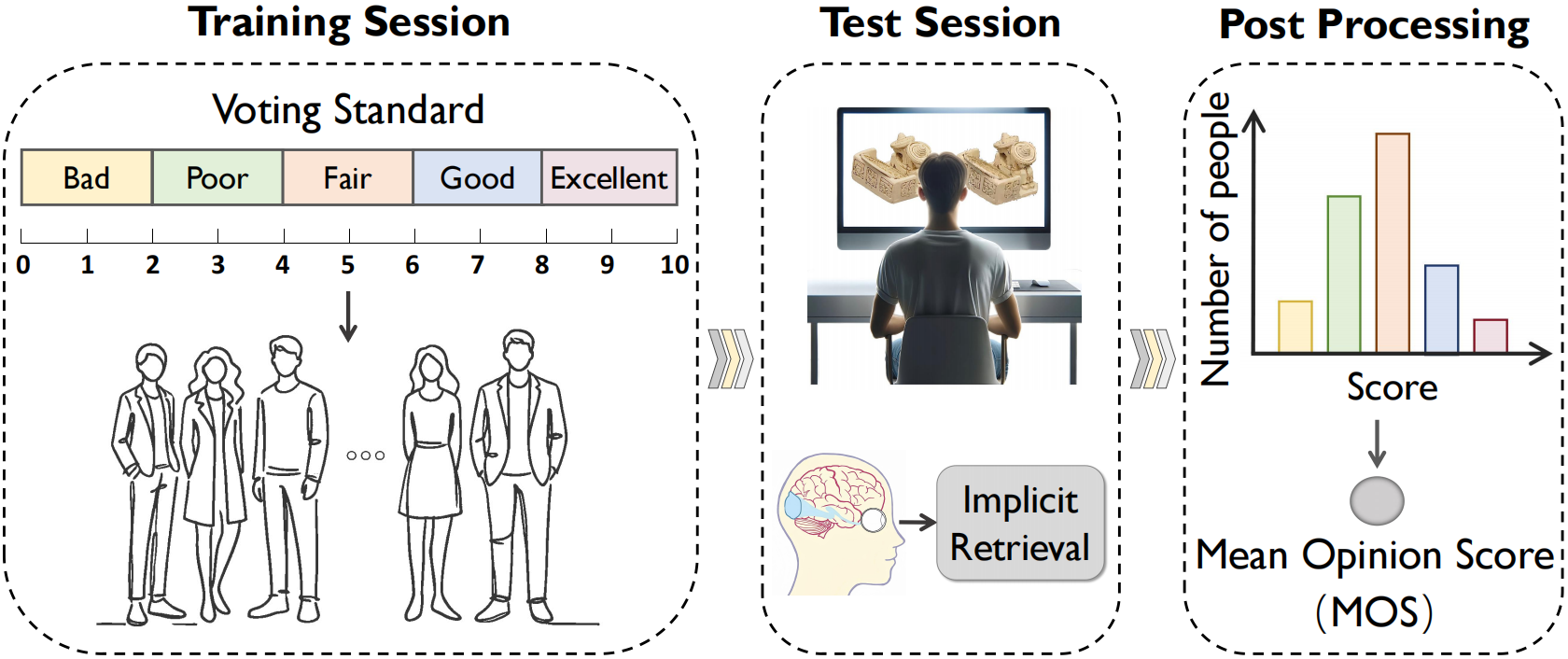}
    \caption{The process of subjective experiment. Participants first undergo a training process to understand the voting standard, then they are able to convert the quality descriptions into quantitative values. Note that multiple subjects may give diverse scores, which forms the score distribution.}
    \label{fig:subjective_experiment}
\end{figure}

In the past years, a variety of deep learning-based NR-PCQA methods have been proposed. These methods can be operated directly on 3D point clouds~\cite{liu2023point, shan2023gpa,tliba2023quality} or 2D projected images~\cite{torlig2018novel,liu2021pqa, mu2023multi}. More recently, another family of methods~\cite{zhang2022mm, xie2023pmbqa} have designed multi-modal architecture to jointly utilize differentiated 2D and 3D information.
These works, regardless of which modality they are based on, try to seek a direct mapping from visual data to quantitative scores. The common strategy is to use well-designed deep neural networks to extract effective quality features from input point clouds, and then regress the features into required scores.

Although promising results have been reported, the prevalent regression-based mapping strategy is somewhat contradictory to the mechanism of subjective evaluation. In fact, when stimulated by distorted samples, people's immediate perception involves intrinsic descriptions such as ``the quality of this sample is good" or ``the distortion is annoying". To acquire quantitative scores that are more convenient in practical applications, participants in subjective experiments need to undergo a training process in which they are informed about the voting standard (\ie, the correspondence between quality/impairment descriptions and quantitative values). As shown in~\Cref{fig:subjective_experiment}, the five-grade categorical scales~\cite{bt500} have been widely used. After the training session, subjects can use quantitative values to represent their judgment by \textit{implicit retrieval}, that is, finding the description that best matches the test sample. We argue that quality descriptions, such as ``excellent" or ``poor", can better simulate human perception. Consequently, we can adopt a two-fold retrieval-based mapping strategy for quality prediction: i) first, find out which language description matches the test sample best; ii) second, represent the language descriptions using quantitative values.

Notably, direct retrieval for a single description may not generate the desired results. In subjective experiments, different subjects may have different understandings and preferences for the same sample, leading to diverse scores. To reduce personal bias, the final MOS is usually obtained by averaging the scores of multiple participants. The distribution of raw subjective opinion scores can provide richer information than MOS, such as individual diversity and uncertainty. Therefore, we further argue that predicting the Opinion Score Distribution (OSD) may be more consistent with subjective evaluation than predicting a scalar score.

Based on the above insights, we develop a novel NR-PCQA method named CLIP-PCQA, which uses a retrieval-based mapping strategy and considers score distributions as final targets. We first extract visual and textual features from point clouds and quality descriptions, where we resort to CLIP~\cite{radford2021learning} that trains a visual encoder and a textual encoder on large-scale (400 million) image-text pairs.
Specifically, to extract visual features, 3D point clouds are projected onto multiple image planes to obtain 2D color and depth maps, which are fed into two separate visual encoders and then simply fused. A contrastive loss is employed to help the model distinguish different samples and viewpoints. For the textual feature extraction, we place different quality-related adjectives (\eg, ``excellent" and ``poor") into a learnable template as prompts and then extract the corresponding features using the frozen textual encoder. 

After the feature extraction, to mimic the retrieval process, we calculate the cosine similarity between the visual feature and multiple textual features corresponding to different quality descriptions. These similarities are then fused into softmax probabilities. We treat the output softmax probabilities as the predicted quality score distribution and use the original opinion score distribution as supervisory information. Finally, by matching each quality description with a certain quantitative value and using the softmax probabilities as weights, we could obtain the quality score of point clouds. Our main contributions are summarized as follows:

\begin{itemize}
\item We propose a novel language-driven PCQA method named CLIP-PCQA, which utilizes the capacity of natural language to understand point cloud quality.
\item We leverage quality descriptions to characterize people's intuitive perception and align with subjective experiments by predicting the OSD. The entire method simultaneously and effectively simulates the mechanisms of HVS and the process of subjective experiments.
\item We conduct comprehensive experiments on multiple benchmarks. Experimental results indicate that CLIP-PCQA achieves superior performance and further analyses reveal the model’s robustness under different settings.
\end{itemize}

\section{Related Works}
\subsection{No-Reference Point Cloud Quality Assessment}
NR-PCQA aims to evaluate the perceptual quality of distorted point clouds without available references. Existing NR methods can be roughly divided into point-based and projection-based methods. Point-based methods use 3D operations on raw 3D points without any transformation. ResSCNN~\cite{liu2023point} pioneered the first NR-PCQA method based on an end-to-end voxel-based sparse CNN. 
GPA-Net~\cite{shan2023gpa} uses a novel graph convolution kernel and a multi-task framework for quality regression, distortion type, and degree predictions. 
3DTA~\cite{zhu20243dta} employs a two-stage sampling method for efficient point cloud representation. Projection-based methods assess the quality based on the 2D projected images from point clouds. PQA-Net~\cite{liu2021pqa} is the first NR-PCQA method based on multi-view feature extraction and fusion. 
D\textsuperscript{3}-PCQA~\cite{liu2023once} establishes an intermediate description domain to bridge perceptual and quality domains.
Fan \etal~\cite{fan2024uncertainty} pioneered a probabilistic architecture through a conditional variational autoencoder.

Some recent works propose to fuse multi-modal information for a comprehensive feature representation of point clouds. 
MM-PCQA~\cite{zhang2022mm} first fuses the texture features from 2D images and geometric information from 3D point clouds.
MMF-PCQM~\cite{wang2023applying} utilizes a dual-branch fusion network and collaborative adversarial learning strategy for multi-modal learning.

\begin{figure*}
    \centering
    \includegraphics[width = 0.92\linewidth]{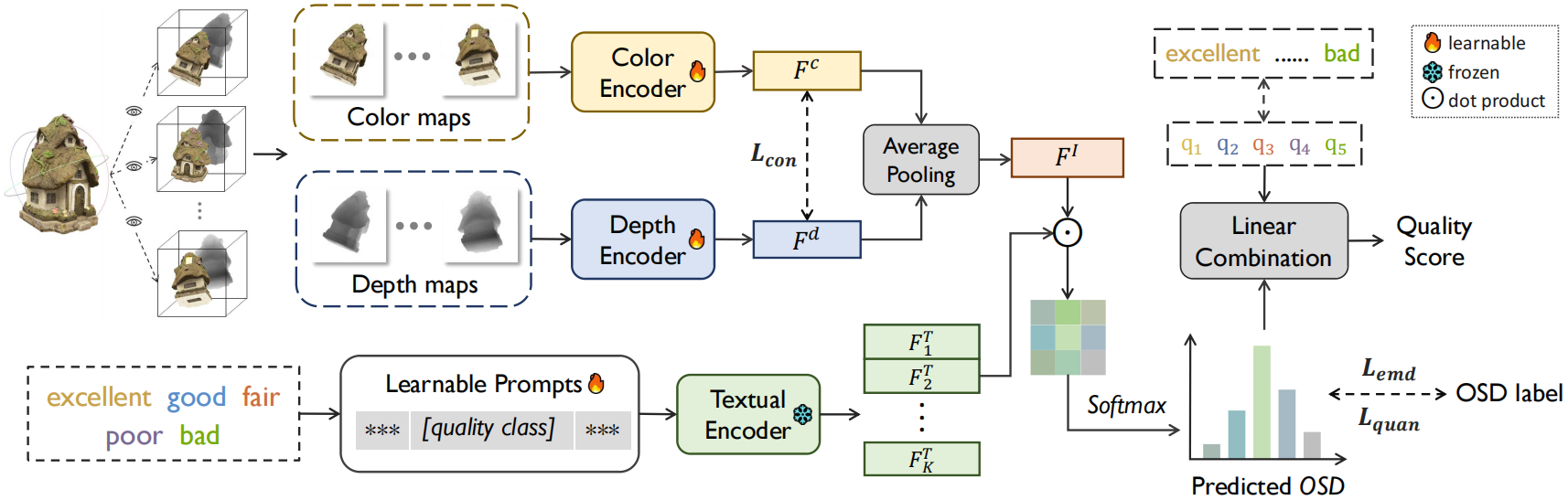}
    \caption{The proposed CLIP-PCQA framework, which includes two main parts: multi-modal feature extraction and vision-language alignment. We use the encoders in CLIP to extract features and then perform vision-language alignment using OSD.}
    \label{fig:framework}
\end{figure*}

\subsection{Vision-Language Learning in Quality Assessment}
Vision-Language (V-L) pre-training has achieved extraordinary success for a wide range of multi-modal tasks. By pretraining large-scale image-text pairs, CLIP~\cite{radford2021learning} has raised a revolution in various vision fields~\cite{he2023clip,gu2021open,luo2022clip4clip,wang2021actionclip,zhang2022pointclip}, which proves the capacity of natural language to understand visual concepts. In the field of quality assessment, some IQA and VQA methods~\cite{wang2023exploring,liu2023ada,wuliao2023towards,wuzhang2023towards,srinath2024learning,agnolucci2024quality} have exploited CLIP for better quality prediction performance. Roughly, these works can be divided into two categories. 

The first category is similar to the framework of CLIP and employs a pair of antonym prompts (positive description and negative description) as textual inputs. CLIP-IQA~\cite{wang2023exploring} and MaxVQA~\cite{wuzhang2023towards} convert quality regression to a binary classification task and regard the softmax results as the quality scores. GRepQ~\cite{srinath2024learning} and QualiCLIP~\cite{agnolucci2024quality} artificially construct image groups with different quality, and then rank similarity or separate higher and lower quality groups based on antonym text prompts.

The second category incorporates the vision-language information from CLIP as one modality in multi-modal tasks to comprehensively assess quality. Ada-DQA~\cite{liu2023ada} constructs a diverse pool that encompasses a broad spectrum of quality-related factors and performs knowledge distillation, the features generated by CLIP and other pretrained models are aggregated. BVQI~\cite{wuliao2023towards} includes the spatial naturalness index, the temporal naturalness index, and the CLIP-based semantic affinity quality index, the three indices are aggregated for final prediction.

The spirit of our work is more similar to the first category. However, previous studies simply employ a pair of antonym prompts (\eg, ``a good photo" and ``a bad photo"), which fails to explore finer-grained descriptions in subjective experiments. To better simulate the process of subjective evaluation, our work merges the quality descriptions in five-grade voting scales into learnable prompts and considers score distributions as final targets. Furthermore, to better exploit the characteristics of point cloud data, we employ an effective contrastive loss to constrain the visual feature extraction.

\section{Proposed Method}
In this study, we aim to develop a language-driven PCQA method that simulates the mechanism of subjective evaluation. The architecture of our network is shown in~\Cref{fig:framework}. The proposed method consists of two main parts: multi-modal feature extraction and vision-language alignment. We will first formulate our PCQA model and then present details of different modules in the following sections.

\subsection{Problem Formulation}
Existing learning-based NR-PCQA approaches typically take point clouds or projected images as input $x$ and seek a direct mapping function to MOS denoted by $Q$. This procedure can be formulated as:
\begin{equation}
    \hat{Q} =  r(\phi(x)),
\label{eq:traditional methods}
\end{equation}
where $\hat{Q}$ indicates the predicted score, $\phi(\cdot)$ denotes the feature extraction operation while $r(\cdot)$ represents a regression head that maps the feature to a scalar value.

As aforementioned, this regression-based mapping strategy does not align well with the subjective assessment. Human beings tend to describe quality using discrete adjectives, and the training session in subjective experiments helps them establish the relation between quality descriptions and quantitative scores. During the test phase, different participants can retrieve different quality descriptions for specific samples based on prior knowledge, and then convert them into quantitative values. To reduce personal limitations and bias, MOS is obtained by averaging the scores of multiple subjects. Therefore, given $K$ quality descriptions $\{y_k\}_{k=1}^K$, we expect to predict the OSD $P=\{p_k\}_{k=1}^K$ for a specific sample, where $p_k$ denotes the probability that subjects choose the $k$-th description. We model the quality evaluation process as follows:
\begin{equation}
\begin{aligned}
    \hat{p}_k &= \mathrm{softmax}(\psi(\phi_1(x),\phi_2(y_k))),
\label{eq:our methods1}
\end{aligned}
\end{equation}
where $\phi_1(\cdot)$ and $\phi_2(\cdot)$ denote visual and textual feature extraction operations, respectively. $\psi(\cdot)$ calculates the similarities between visual features and textual features of different quality descriptions, which are then transformed into probabilities using the softmax operation. Note that we directly derive the probabilities by comparing visual and textual features and regard them as the uncertainty in subjective experiments. As a result, the process of subjective evaluation can be approximated as sampling from this distribution. 

By employing the novel quality prediction strategy, we actually divide the PCQA task into two steps: multi-modal feature extraction and vision-language alignment. We use the visual and textual encoders in CLIP to extract multi-modal features and then perform vision-language alignment using the OSD. A contrastive loss and learnable prompts are introduced to enhance the feature extraction. The proposed quality prediction procedure is more consistent with the way how humans perceive point cloud quality and we will present the detailed pipeline of our method in the following.

\subsection{Preprocessing}
CLIP has a powerful capability to align visual data with language concepts. However, its visual encoder is pre-trained on 2D images while point clouds are sets of unordered points in 3D space. To convert point clouds into CLIP-accessible inputs, we project the 3D point clouds into multi-view images. To reduce information loss during the projection process, we obtain both color and depth maps from $M$ viewpoints simultaneously, which is clearly shown in~\Cref{fig:framework}. 
Color maps offer texture details about point clouds, while depth maps provide geometric information about the shape, structure, and point distribution. Given a point cloud $X$, We denote its multi-view color and depth maps by $\{X_m^c\in \mathbb{R}^{H\times W\times3}|_{m=1}^M\}$ and $\{X_m^d\in \mathbb{R}^{H\times W}|_{m=1}^M\}$ respectively, where $m$ denotes the $m$-th viewpoint.

\begin{figure}[t]
    \centering
    \includegraphics[width=0.95\linewidth]{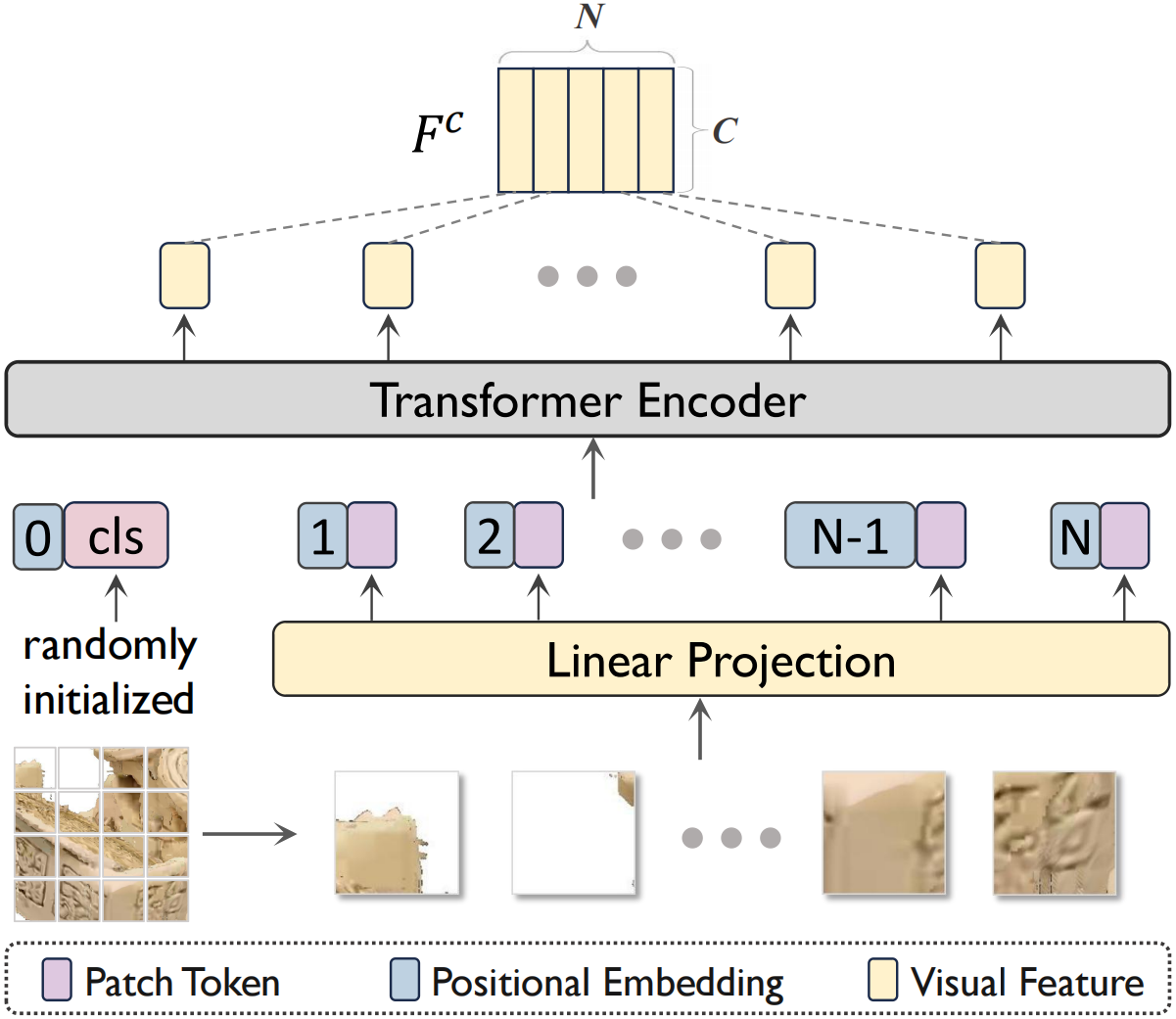}
    \caption{Illustration of the visual encoder architecture.
    }
    \label{fig:vit}
\end{figure}

\subsection{Multi-Modal Feature Extraction}
\subsubsection{Visual Feature Extraction}

We first intend to obtain visual features from point clouds. Note that previous studies usually freeze the visual encoder of CLIP to perform zero-shot prediction. However, features obtained from the frozen encoder can lead to a large performance gap for PCQA task because of the differences between point cloud projections and natural images. Therefore, we choose to finetune the visual encoder of CLIP in the training process. 

The visual encoder in CLIP is capable of processing depth maps~\cite{zhang2022pointclip}. Given the distinct characteristics of color and depth maps, we employ two pretrained Vision Transformer (ViT) from CLIP, denoted by $\phi_{c}$ and $\phi_{d}$, which are independently updated during the training. 
To make the projected images compatible with the ViT, for $X^c$ or $X^d$ from an arbitrary perspective, we partition each image into $N$ non-overlapping patches and feed the patches into the encoders. More specifically, we first flatten the patches and map them to $C$-dimension patch embeddings using a trainable linear projection. Together with a randomly initialized class token, the token sequence can be denoted as $z_0=\{f_{cls},f_1,...,f_{N}\}\in\mathbb{R}^{{(N+1)}\times{C}}$. 

Then the position embeddings are added to all tokens to retain positional information. Subsequently, we input these tokens into the transformer consisting of $L$ blocks. For instance, the procedure in the $l$-th block can be formulated as:
\begin{equation}
    z_l=\mathrm{MLP}(\mathrm{LN}(z'_l))+z'_l, z'_l=\mathrm{MSA}(\mathrm{LN}(z_{l-1}))+z_{l-1},
\end{equation}
where LN stands for layer normalization, MLP denotes the multi-layer perceptron and MSA denotes the multi-head self-attention.

Finally, we could obtain the last-layer output of all tokens $z_L\in\mathbb{R}^{{(N+1)}\times{C}}$. 
To better account for multiple local distortions across the entire image, we use the feature map of all patches with the dimension of $N\times C$ instead of the class token as in CLIP. 

We illustrate the whole feature extraction process for an arbitrary color map in~\Cref{fig:vit}. Mathematically, given the color map and depth map of the $m$-th viewpoint (\ie, $X_m^c$ and $X_m^d$), the output features can be formulated as: 
\begin{equation}
\label{eq:feature extraction}
    \begin{aligned} 
    \begin{array}{cc}
    & {F}^{c}_m = \{f_{n,m}^c\}_{n=1}^N= \phi_{c}(X^c_m)\in\mathbb{R}^{{N}\times{C}}, \\[2mm]
    & {F}^{d}_m = \{f_{n,m}^d\}_{n=1}^N =\phi_{d}(X^d_m)\in\mathbb{R}^{{N}\times{C}},
    \end{array}
    \end{aligned} 
\end{equation}
where $f_{n,m}^c\in\mathbb{R}^C$ (or $f_{n,m}^d\in\mathbb{R}^C$) denote the output token corresponding to the $n$-th color or depth patch of the $m$-th viewpoint. To merge the information of different patches and viewpoints, the final visual feature is calculated as follows:
\begin{equation}
{F}^{I} = \frac{1}{2MN}\sum_{n=1}^N\sum_{m=1}^M(f_{n,m}^c+f_{n,m}^d) \in\mathbb{R}^{C}.
\end{equation}

To enable the model to distinguish different samples and viewpoints, we introduce a contrastive scheme. Specifically, the color and depth maps from the same view of a specific sample are considered positive pairs, while others are treated as negative ones. We first flattened the features in~\Cref{eq:feature extraction} into $\Tilde{F}^{c}_m\in\mathbb{R}^{{NC}}$ and $\Tilde{F}^{d}_m\in\mathbb{R}^{{NC}}$, respectively. Then the contrastive loss is calculated as follows:
\begin{small}
\begin{equation}
\begin{aligned}
{L}_{con} = \dfrac{1}{2BM}& \sum^{B}_{b=1}\sum^{M}_{m=1} \big( l_{con}(\Tilde{F}^{c}_{m,b},\Tilde{F}^{d}_{m,b})
+ l_{con}(\Tilde{F}^{d}_{m,b},\Tilde{F}^{c}_{m,b}) \big), \\
{l}_{con}(i,&j) = -\mathrm{log} \frac{\exp(\mathrm{cos}(i, j)/\tau_1)}{\sum_{k=1}^{BM} 1_{[k \neq i]}\exp(\mathrm{cos}(i, k)/\tau_1)},
\end{aligned}
\label{eq:contrastiveloss}
\end{equation}
\end{small}
where $\Tilde{F}^{c}_{m,b}$ (or $\Tilde{F}^{d}_{m,b}$) denotes the flattened color (or depth) feature corresponding to the $m$-th viewpoint of the $b$-th sample in a mini-batch.
$B$ is the batch size, $1_{[k \neq i]}$ is an indicator function evaluating to $1$ if $k \neq i$, $\mathrm{cos}(\cdot)$ calculates the cosine similarity between two features and $\tau_1$ is temperature. Obviously, the contrastive loss minimizes the distance between color and depth features from the same view of a specific sample, while maximizing the distance of other pairs, which helps the network better characterize the point cloud data.

\subsubsection{Textual Feature Extraction}
The textual encoder in CLIP is capable of projecting textual descriptions into the embedding space close to the features of the paired images. To take advantage of this semantic analysis capability, we convert different quality-related adjectives into prompts and freeze the textual encoder to extract the corresponding features.

Concretely, to simulate the training session in subjective experiments, we adopt the widely used five quality-related adjectives from ITU-R BT.500~\cite{bt500} (\ie, ``excellent'', ``good'', ``fair'', ``poor'' and ``bad'') as $K$ quality level tokens and obtain their embeddings as $A={\{A_k\}}_{k=1}^K$. Considering that fixed prompts (\eg, ``the
quality of this image is \textit{excellent}" or ``an image of \textit{excellent} quality") may restrict the space of model optimization, we introduce the unified learnable prompt version following CoOp~\cite{zhou2022learning}. More specifically, we insert the quality level tokens into the middle of the template sentence to acquire the quality-related prompts $Y={\{Y_k\}}_{k=1}^K$, that is
\begin{equation}
Y_k={[V]}_1...{[V]}_{W/2}[A_k]{[V]}_{W/2+1}...{[V]}_W,
\label{eq:learnableprompt}
\end{equation}
where each $[V]$ is a learnable context token with the same dimension as word embedding, and $W$ is a hyper-parameter specifying the number of context tokens.

Subsequently, these prompts will pass through the frozen textual encoder $\phi_t$ to obtain a list of features as:
\begin{equation}
F^T=\{{F}_{k}^{T}\}^{K}_{k=1}\in\mathbb{R}^{{K}\times{C}},  F^T_k= \phi_t(Y_k),
\label{eq:textualencoder}
\end{equation}
where $F_k^T$ denotes the textual feature corresponding to the $k$-th quality description.

\subsection{Vision-Language Alignment}

The aforementioned analysis argues that participants in subjective experiments tend to first match the distorted samples with quality descriptions and then represent the quality descriptions using quantitative values. To mimic such a process, we first calculate the cosine similarities between image features and multiple textual features corresponding to different quality descriptions:
\begin{equation}
\pi_k = \dfrac{{F}^{I}\cdot({{F}^T_k)^{\bf{T}}}}{{\lVert{{F}^{I}}\rVert}{\lVert{{F}^{T}_k}\rVert}}.
\end{equation}
The larger the similarity, the more likely the corresponding quality description is to be retrieved. Given the existence of personal limitations and bias in subjective experiments, we further transform the similarities into probabilities. Concretely, we leverage a softmax function to convert the similarity values into probabilities to acquire the predicted OSD $\hat{P}=\{\hat{p}_k\}_{k=1}^K$ as:
\begin{equation}
\hat{p}_k=\dfrac{\exp({{\pi}_{k}})}{\sum_{i=1}^{K}\exp({{\pi}_{i}})},
\end{equation}
where we treat the output softmax probabilities as the predicted quality score distribution. Generally, we regard that the derived probabilities indicate the uncertainty in subjective experiments. As a result, the process of subjective evaluation can be approximated as sampling from this distribution. We expect $\hat{P}$ to approximate the ground truth OSD during the training phase.

In practical application scenarios, we prefer the model to output an overall quality score for a given point cloud sample, which benefits the quality-oriented optimization. Therefore, in the test phase, we quantify the $K$ quality-related descriptions into certain quantitative values. For example, [``excellent'', ``good'', ``fair'', ``poor'' and ``bad''] are projected to quality scores $q =\{q_k\}_{k=1}^K=[5, 4, 3, 2, 1]$ for databases with MOS = 5 as the best
quality. Using the softmax probabilities as weights, the final predicted score $\hat{Q}$ is calculated by a linear combination of the predicted OSD and the score values of different quality levels:
\begin{equation}
    \hat{Q} = \sum_{k=1}^{K}\hat{p}_{k}*q_k,\text{ for } k = 1,...,K.
\label{eq:calmos}
\end{equation}

\subsection{Loss Function}
Based on the existing findings~\cite{hossfeld2015qoe}, subjective uncertainty can be expressed by the Cumulative Distribution Function (CDF). Therefore, to constrain the network optimization, we leverage the Earth Mover’s Distance (EMD) following~\cite{hou2016squared} to measure the discrepancy between the predicted OSD $\hat{P}=\{\hat{p}_k\}_{k=1}^K$ and the ground truth OSD $P=\{p_l\}_{l=1}^L$ as:
\begin{equation}
L_{emd}=\left(\dfrac{1}{K}\sum^{K}_{k=1}{|\mathrm{CDF}_{\hat{P}}(q_k)-\mathrm{CDF}_P(q_k)|}^2\right)^{{1}\slash{2}},
\end{equation}
where $L$ represents the number of score options available to participants in the original scoring. $\mathrm{CDF}(\cdot)$  returns the cumulative probability and can be formulated as:
\begin{equation}
\mathrm{CDF}_P(q_k)=\mathrm{Pr}(Q\leq q_k).
\end{equation}

Following~\cite{gao2022image}, we further introduce a quantile-based loss function. Quantiles refer to numerical points that divide the probability distribution range of a random variable into several equal parts, and $\theta$-quantile denotes the specific score whose CDF equals $\theta$. Since $\mathrm{CDF}_{\hat{P}}$ and $\mathrm{CDF}_{P}$ are discrete, we first use a linear interpolation algorithm to convert them into continuous distributions. Then
the quantile-based loss function can be calculated as:
\begin{equation}
L_{quan}=\dfrac{1}{J}\sum^{J}_{j=1}{|s_{\hat{P}}(\theta_j)-s_P(\theta_j)|},
\end{equation} 
where $\theta_j\in\mathbb{\boldsymbol{\theta}}=[\theta_1,\theta_2,...,\theta_J]$ and J is the number of quantiles we calculated. $s_{\hat{P}}(\theta_j)$ and $s_P(\theta_j)$ represent the predicted $\theta_j$-quantile and the ground-truth $\theta_j$-quantile. 

The final loss function is the combination of three losses:
\begin{equation}
Loss = L_{emd} + \alpha L_{quan} + \beta L_{con},
\end{equation}
where $\alpha$ and $\beta$ are hyper-parameters used to control the scale of the three loss functions.

\begin{table*}[th]\small
\begin{center}
\renewcommand{\tabcolsep}{1.4mm}
\centering
\scalebox{0.87}{
\begin{tabular}{ccp{0.07\textwidth}cccccccc}
\toprule
\toprule
   \multirow{2}{*}{Type} & \multirow{2}{*}{Method} & \multicolumn{3}{c}{SJTU-PCQA~\cite{yang2020predicting}} & \multicolumn{3}{c}{LS-PCQA~\cite{liu2023point}} & \multicolumn{3}{c}{BASICS~\cite{ak2024basics}} \\ 
\cmidrule(l){3-5} \cmidrule(l){6-8} \cmidrule(l){9-11}
& & PLCC$\uparrow$ & SRCC$\uparrow$ & RMSE$\downarrow$ & PLCC$\uparrow$ & SRCC$\uparrow$ & RMSE$\downarrow$ & PLCC$\uparrow$ & SRCC$\uparrow$ & RMSE$\downarrow$ \\ \midrule\midrule
\multirow{8}{*}{FR}
& MSE-p2po~\cite{mekuria2016evaluation} & 0.896 & 0.810 & 1.046 & 0.427 & 0.301 & 0.744 & 0.849 & 0.774 & 0.559 \\
& MSE-p2pl~\cite{mekuria2016evaluation} & 0.783 & 0.696 & 1.471 & 0.454 & 0.286 & 0.734 & 0.895 & \textcolor{blue}{0.836} & 0.468 \\
& PSNR\textsubscript{yuv}~\cite{torlig2018novel} & 0.764 & 0.762 & 1.513 & 0.527 & 0.482 & 0.699 & 0.617 & 0.592 & 0.835 \\
& PCQM~\cite{meynet2020pcqm} & 0.836 & 0.874 & 2.378 & 0.208 & 0.426 & 0.789 & 0.891 & 0.808 & 0.478 \\
& PointSSIM~\cite{alexiou2020towards} & 0.751 & 0.725 & 1.539 & 0.225 & 0.164 & 0.804 & 0.780 & 0.731 & 0.657 \\
& MPED~\cite{yang2022mped} & 0.903 & 0.900 & 1.000 & 0.613 & 0.609 & 0.646 & 0.807 & 0.712 & 0.615 \\
& GraphSIM~\cite{yang2020inferring} & 0.896 & 0.874 & 1.040 & 0.358 & 0.331 & 0.767 & 0.895 & 0.813 & 0.465 \\
& TCDM~\cite{zhang2023tcdm} & 0.952 & 0.929 & 0.713 & 0.438 & 0.413 & 0.739 & 0.874 & 0.757 & 0.505 \\
\midrule
\multirow{6}{*}{NR}
&PQA-Net~\cite{liu2021pqa} & 0.898 & 0.875 & 1.340 & 0.607 & 0.595 & 0.697 & 0.655 & 0.387 & 0.797 \\
&ResSCNN~\cite{liu2023point} & 0.889 & 0.880 & 0.878 & 0.648 & 0.620 & 0.615 & 0.391 & 0.352 & 0.975 \\
&MM-PCQA~\cite{zhang2022mm} & 0.939 & 0.910 & 0.805 & 0.644 & 0.605 & 0.621 & 0.793 & 0.738 & 0.628 \\
&GMS-3DQA~\cite{zhang2024gms} & 0.916 & 0.886 & 0.931 & \textcolor{blue}{0.666} & \textcolor{blue}{0.645} & \textcolor{blue}{0.606} & 0.895 & 0.807 & 0.472 \\
&3DTA~\cite{zhu20243dta} &  \textcolor{blue}{0.953} & \textcolor{blue}{0.931} & \textcolor{blue}{0.706} & 0.613 & 0.604 & 0.645 & \textcolor{blue}{0.907} & 0.825 & \textcolor{blue}{0.437} \\
&\textbf{CLIP-PCQA (ours)} &\bf\textcolor{red}{0.956} &\bf\textcolor{red}{0.936} &\bf\textcolor{red}{0.693} &\bf\textcolor{red}{0.755} &\bf\textcolor{red}{0.736} &\bf\textcolor{red}{0.533} &\bf\textcolor{red}{0.932} &\bf\textcolor{red}{0.872} &\bf\textcolor{red}{0.382} \\
\bottomrule
\bottomrule
\end{tabular}
}
\caption{Benchmarking results of SOTA methods on the SJTU-PCQA, LS-PCQA and BASICS databases. $``\uparrow"/``\downarrow"$ indicates that larger/smaller is better. Best in {\bf\textcolor{red}{red}} and the second in \textnormal{{\textcolor{blue}{blue}}.}
}
\label{tab:experiment}
\end{center}
\end{table*}

\section{Experiments}
\subsection{Databases and Evaluation Metrics}
\textbf{Databases.} To illustrate the effectiveness of our method, we employ three benchmarks with available raw opinion scores: SJTU-PCQA~\cite{yang2020predicting}, LS-PCQA Part I~\cite{liu2023point} and BASICS~\cite{ak2024basics}. SJTU-PCQA includes 9 native point clouds and 378 distorted point clouds disturbed by 7 types of distortion under 6 levels. LS-PCQA comprises 85 native point clouds, and a total of 930 labeled point clouds generated by 31 distortion types. BASICS consists of 1494 distorted point clouds impaired by 4 types of compression, which are generated from 75 references.

\textbf{Evaluation Metrics.}
Three widely adopted evaluation metrics are employed to quantify the level of agreement between the predicted quality scores and the MOSs: Pearson Linear Correlation Coefficient (PLCC), Spearman Rank Correlation Coefficient (SRCC), and Root Mean Squared Error (RMSE). Since there exists a misalignment in the value range between the predicted scores and the MOSs, we introduce a common four-parameter logistic function~\cite{antkowiak2000final} to align their ranges.

\begin{table}[t]\small
\begin{center}
\renewcommand{\tabcolsep}{1.8mm}
\centering
\scalebox{0.88}{%
\begin{tabular}{ccccc}
\toprule
\toprule
Train on & \multicolumn{2}{c}{SJTU-PCQA} & \multicolumn{2}{c}{LS-PCQA} \\
\cmidrule(l){1-1} \cmidrule(l){2-3} \cmidrule(l){4-5} 
Test on & LS-PCQA & BASICS & SJTU-PCQA & BASICS \\ 
\midrule
ResSCNN & 0.300 & 0.092 & 0.472 & 0.078 \\
MM-PCQA & 0.150 & 0.198 & 0.728 & 0.445 \\
GMS-3DQA & \bf\textcolor{red}{0.366} & 0.575 & \textcolor{blue}{0.811} & \textcolor{blue}{0.457} \\
3DTA & 0.288 & \textcolor{blue}{0.633} & 0.671 & 0.448 \\
\textbf{CLIP-PCQA} & \textcolor{blue}{0.326} & \bf\textcolor{red}{0.685} & \bf\textcolor{red}{0.873}  & \bf\textcolor{red}{0.570} \\

\bottomrule
\bottomrule
\end{tabular}
}
\caption{Evaluation of cross-database generalization. The training and test are all performed on the complete database. PLCC are reported, best in {\bf\textcolor{red}{red}} and the second in \textnormal{{\textcolor{blue}{blue}}}.
}
\label{tab:Cross-database evaluation}
\end{center}
\end{table}

\subsection{Implementation Details}

\textbf{Database Split.} We partition the databases according to content (reference point clouds) and k-fold cross-validation is used for training. Specifically, 9-fold cross-validation is applied for SJTU-PCQA following~\cite{zhang2022mm}, and we adopt a 5-fold cross-validation both for LS-PCQA and BASICS. For each fold, the test performance with minimal training loss is recorded and the average result across all folds is recorded to alleviate randomness. 

\textbf{Network Details.} We use the basic version of Vision Transformer~\cite{dosovitskiy2020image} with 16 × 16 patch embeddings (namely ViT-B/16) in CLIP-Visual as our visual encoders. The textual encoder is also the pre-trained transformer in CLIP-Text. 
The number of projection views $M=6$ and the images are randomly cropped into 224×224×3 as inputs. We set the number of context tokens $W$ as 16. For the loss function, we set $\boldsymbol{\theta}=[0.25,0.50,0.75]$. $\alpha$ is set to $1/K$ and $\beta$ is set to 0.08.

\textbf{Training Strategy.} The initial learning rate is set as 4e-6 and the model is trained for 50 epochs with a default batch size of 16. We use the Adam optimizer~\cite{kingma2014adam} with a weight decay of 1e-4.
Depending on the raw score ranges of different databases, we evenly divide them into five thresholds as the quantitative values $q$. For example, we set $q=[5,4,3,2,1]$ for LS-PCQA and $q=[10,8,6,4,2]$ for SJTU-PCQA, respectively. 

\begin{figure}[t]
\centering
\includegraphics[width=1\linewidth]{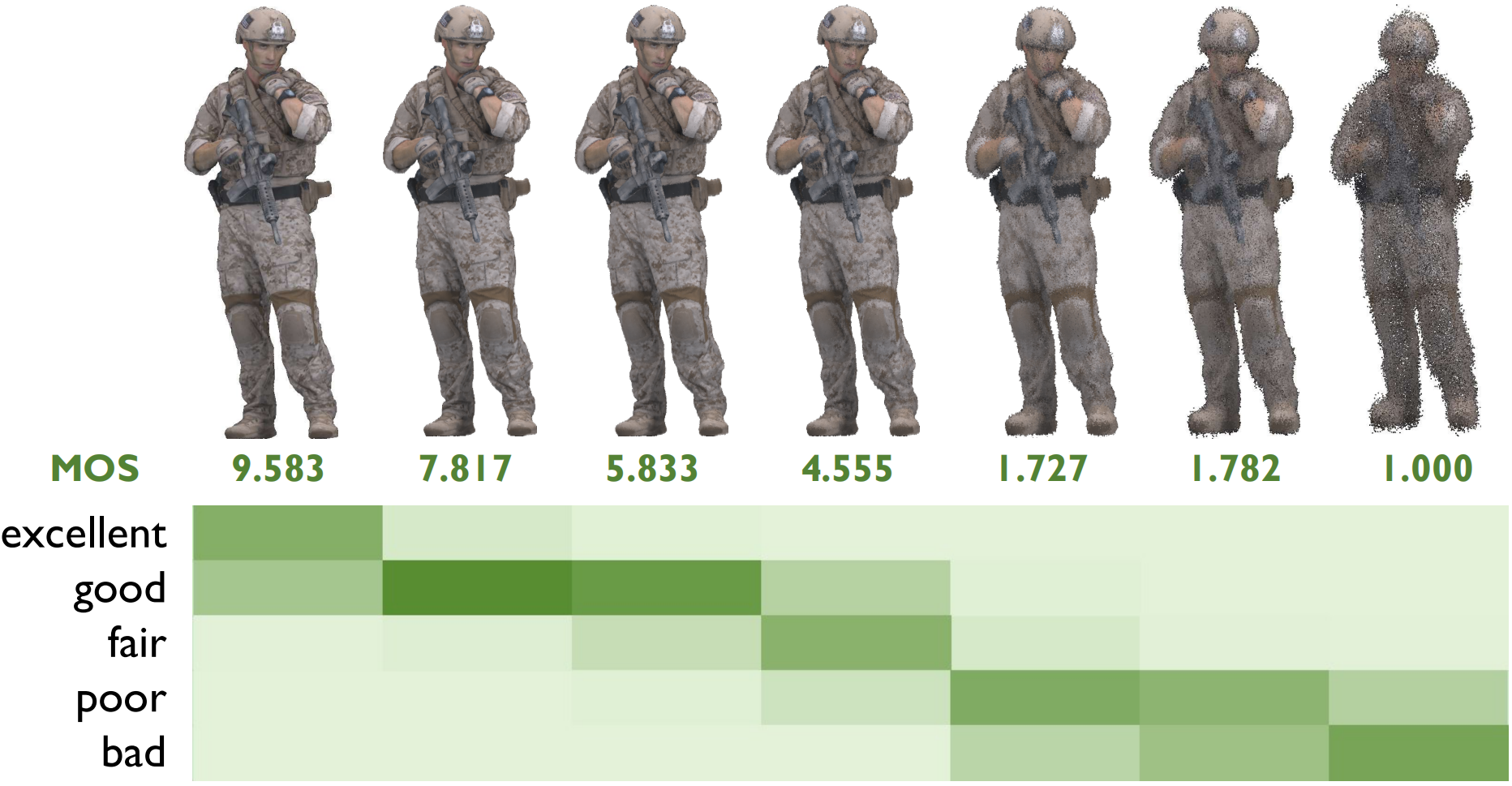}
\caption{Weight matrice visualization. Darker colors indicate higher values, which mean greater correlations.}
\label{fig:visualization}
\end{figure}

\subsection{Performance Comparison}

\textbf{Overall Performance.} For the sake of comprehensively investigating the prediction performance of CLIP-PCQA, 13 SOTA methods are selected for comparison, including 8 FR-PCQA methods and 5 NR-PCQA methods. We strictly retrain the baselines with available code, using the same database split to ensure a fair comparison. The experimental results are illustrated in~\Cref{tab:experiment}. Benefiting from data-driven deep learning approaches, many existing NR methods outperform traditional FR methods. By introducing vision-language alignment and distribution prediction, our method CLIP-PCQA yields the best performance across three databases. Notably, our method significantly surpasses the second-best results on the LS-PCQA (exceeding 0.089 in PLCC and 0.091 in SRCC). Furthermore, while the performance of other methods varies greatly across different databases, our approach is robust to databases of different domains or various point cloud contents and distortions.  

\textbf{Cross-database Evaluation.}
To test the generalizability of our method, we further perform cross-database evaluation. We train the compared methods on SJTU-PCQA and LS-PCQA, and test the trained model on the other databases. The results are reported in~\Cref{tab:Cross-database evaluation}, where our CLIP-PCQA achieves the best generalization performance in most cases. Since the contents are limited in SJTU-PCQA while abundant in LS-PCQA, almost all methods perform poorly when trained on SJTU-PCQA and tested on LS-PCQA. It is worth noting that our method exceeds the other counterparts by a noticeable margin when trained on LS-PCQA and generalized to SJTU-PCQA. Our cross-database result of 0.873 even presents competitive performance compared to the intra-database results in~\Cref{tab:experiment}, which indicates that our method learned more general quality-related features when trained on databases with richer contents and distortions.

\begin{figure}[t]
    \centering
    \begin{subfigure}[b]{0.47\linewidth}
        \centering
        \includegraphics[width=0.97\linewidth]{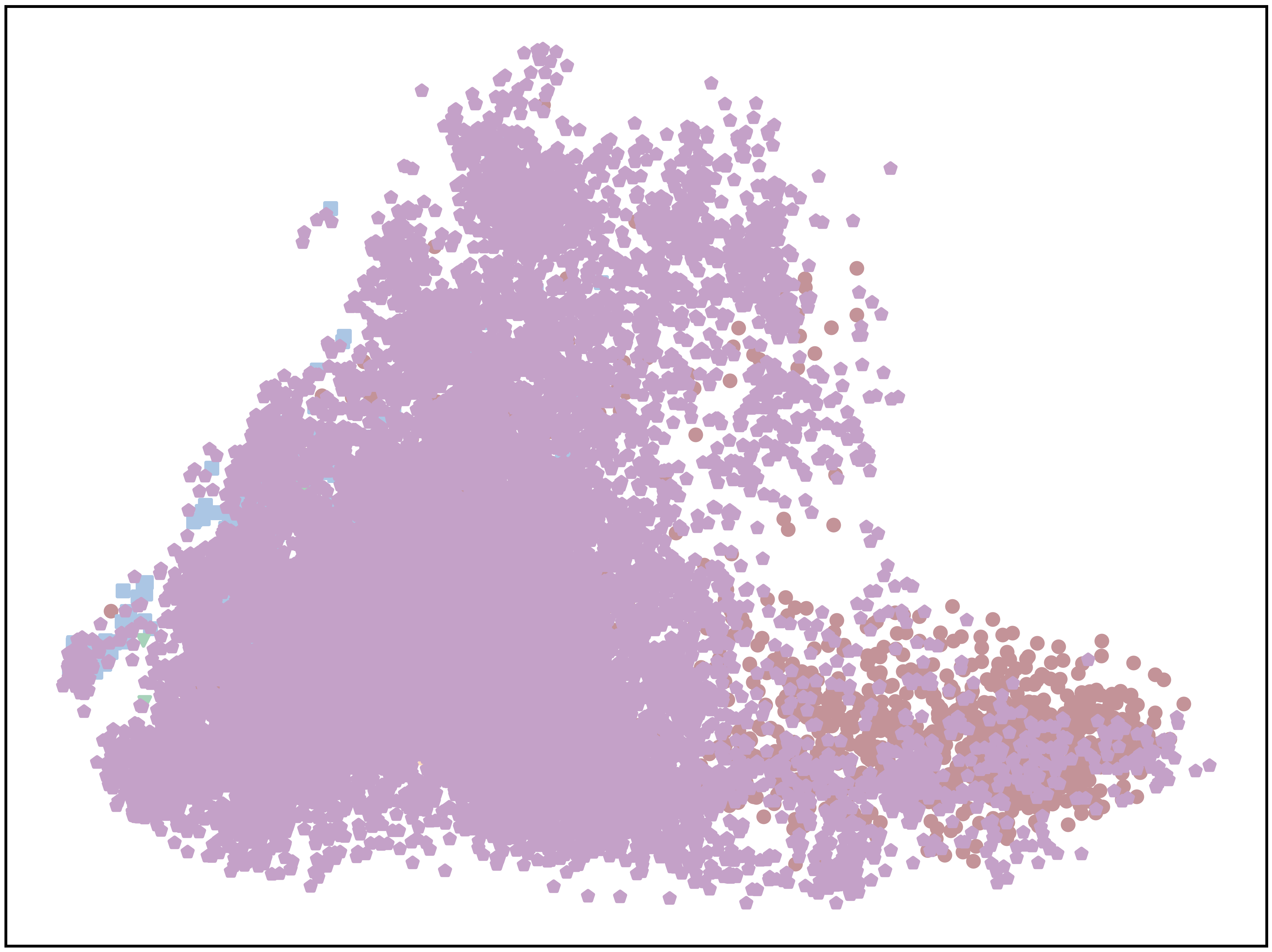}
        \caption{Initialization}
    \end{subfigure}
    \hfill
    \begin{subfigure}[b]{0.47\linewidth}
        \centering
        \includegraphics[width=0.97\linewidth]{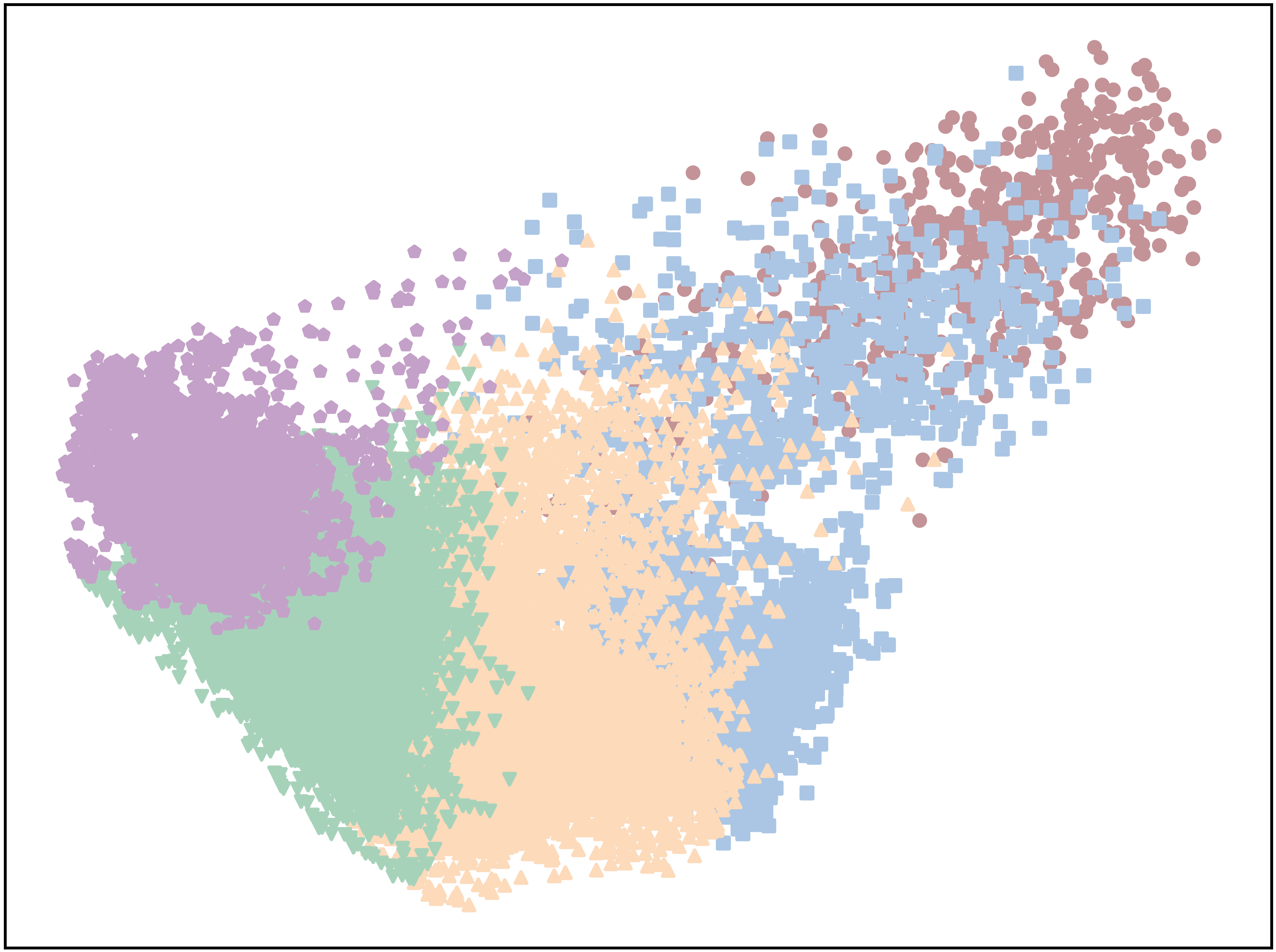}
        \caption{Using trained model}
    \end{subfigure}
    \begin{subfigure}[b]{0.9\linewidth}
        \centering
        \includegraphics[width=0.97\linewidth]{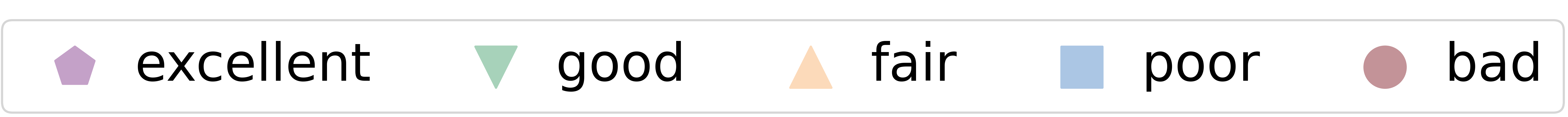}
    \end{subfigure}
    \caption{PCA of the visual features on complete LS-PCQA.}
    \label{fig:classification}
\end{figure}

\begin{table}[t]\small
\begin{center}
  \scalebox{0.95}{%
  \begin{tabular}{cccccc}
    \toprule
    Index & Modality & PLCC$\uparrow$ & SRCC$\uparrow$ & RMSE$\downarrow$\\
    \midrule
    (1) &Color + Text & 0.738 & 0.716 & 0.550 \\
    (2) &Depth + Text & 0.536 & 0.395 & 0.694 \\
    (3) &Color + Depth & 0.688 & 0.652 & 0.585 \\
    (4) &$\textbf{Color + Depth + Text}$ & \textbf{0.755} & \textbf{0.736} & \textbf{0.533} \\
  \bottomrule
\end{tabular}
}
\caption{Contributions of different modalities on LS-PCQA.}
\label{tab:different modalities}
\end{center}
\end{table}

\subsection{Visualization}
To validate that our model has learned the correspondence between the quality-related descriptions and the distortion degrees, we provide qualitative results in this section.

First, we visualize the output softmax probabilities of different quality levels in~\Cref{fig:visualization}. It can be observed that samples with more severe distortions exhibit a stronger correlation with text descriptions indicating lower quality, which confirms that quality information is indeed encapsulated within the textual features.

In addition, we use Principal Component Analysis (PCA) to embed visual features into 2D representation space, which is illustrated in~\Cref{fig:classification}. Specifically, to intuitively demonstrate the clustering results of the samples with different quality, we assign each sample to the quality level category with the highest probability in the softmax output. Since PCQA is essentially a continuous regression task rather than a classification task, it is nearly impossible to distinctly classify all samples into five categories. From~\Cref{fig:classification}, we can observe that distorted samples within the same category are clustered in the feature space. Moreover, the adjacent categories exhibit continuity, indicating the lack of clear boundaries between different quality levels.

\begin{table}[t]\small
\begin{center}
  \scalebox{0.95}{%
  \begin{tabular}{cccccc}
    \toprule
    Type & Position & PLCC$\uparrow$ & SRCC$\uparrow$ & RMSE$\downarrow$\\
    \midrule
    fixed & -  & 0.729 & 0.696 & 0.559 \\
    \multirow{3}{*}{\textbf{learnable}}
    & begin & 0.728 & 0.702 & 0.559 \\
    & \textbf{middle} & \textbf{0.755} & \textbf{0.736} & \textbf{0.533} \\
    & end & 0.742 & 0.719 & 0.548 \\
  \bottomrule
\end{tabular}
}
\caption{Ablation on the prompt type and the position of quality level token on LS-PCQA.}
\label{tab:class token position}
\end{center}
\end{table}

\begin{table}[t]\small
\begin{center}
  \scalebox{0.95}{%
  \begin{tabular}{cccccc}
    \toprule
    $L_{emd}$ & $L_{quan}$ & $L_{con}$ & PLCC$\uparrow$ & SRCC$\uparrow$ & RMSE$\downarrow$\\
    \midrule
    \checkmark & & & 0.734 & 0.717 & 0.553 \\
    & \checkmark & & 0.649 & 0.606 & 0.592 \\
    \checkmark & \checkmark &  & 0.743 & 0.721 & 0.548 \\
    \checkmark &  & \checkmark & 0.745 & 0.722 & 0.542 \\
    \checkmark & \checkmark & \checkmark & \textbf{0.755} & \textbf{0.736} & \textbf{0.533} \\

  \bottomrule
\end{tabular}
}
\caption{Ablation of the loss function on LS-PCQA.}
\label{tab:loss_function}
\end{center}
\end{table}

\subsection{Ablation Studies}

\textbf{Contributions of Modalities.} We conduct
experiments to validate the effectiveness of three types of input and report the results in~\Cref{tab:different modalities}. Seeing (1), (2) and (4), combining color and depth maps enhances the model's ability to gain better quality representations. Since color maps include more visual information, the performance deteriorates when only depth maps are used. Moreover, we assess the performance of removing the text branch and using a two-fold MLP for regression on visual features in (3). Comparing (3) and (4), we can conclude that incorporating language significantly improves performance.

\textbf{Prompt Design.} 
To validate the effectiveness of learnable prompts, we perform experiments with fixed prompt ``The quality of this image is [CLASS]''. Additionally, we test the performance with different positions of the class token and report the results in~\Cref{tab:class token position}. We can observe that appropriate learnable prompts outperform fixed prompts and alleviate the need for meticulous prompt design. Moreover, positioning the quality level token in the middle performs best, where bidirectional contextual information helps the model better understand the semantics of the class token. 

\textbf{Loss Function.} We explore how the three loss functions contribute to our model in~\Cref{tab:loss_function}. It can be seen that the predictive performance using only EMD or quantile-based loss is inferior to using both. Additionally, the contrastive loss $L_{con}$ helps the model further improve performance.

\section{Conclusion}
In this paper, to simulate the HVS mechanism and reflect subjective diversity and uncertainty, we propose a novel language-driven method CLIP-PCQA, which predicts quality according to discrete quality descriptions and quality score distribution. Experiments demonstrate that our method yields the SOTA performance, and we believe our study could shed light on future research.

\section*{Acknowledgments}
This paper is supported in part by National Natural Science Foundation of China (62371290, U20A20185), the Fundamental Research Funds for the Central Universities of China, and 111 project (BP0719010). The corresponding author is Yiling Xu(e-mail: yl.xu@sjtu.edu.cn).

\bibliography{aaai25}

\newpage

\section*{Appendix}

\subsection{Five-grade Categorical Scales}
Participants in subjective experiments need to undergo a training process in which they are informed about the voting standard. As shown in \Cref{tab:quality and impairment scales}, five-grade categorical scales ~\cite{bt500} have been widely used for training. After the training session, subjects can use quantitative values to represent their judgment by implicit retrieval, that is, finding the description that best matches the test sample.

\begin{table}[H]
\begin{center}
  \begin{tabular}{ccc}
    \toprule
    Level & Quality & Impairment\\
    \midrule
    5 & Excellent & Imperceptible \\
    4 & Good & Perceptible, but not annoying \\
    3 & Fair & Slightly annoying \\
    2 & Poor & Annoying \\
    1 & Bad & Very annoying \\
  \bottomrule
\end{tabular}
\caption{ITU-R five-grade quality and impairment scales.}
\label{tab:quality and impairment scales}
\end{center}
\end{table}

\subsection{Additional Ablation Studies}
\subsubsection{The choice of visual encoder}
We compare different image backbones of CLIP~\cite{radford2021learning} from ResNet~\cite{he2016deep} to Vision Transformer~\cite{dosovitskiy2020image} and present the quantitative results in \Cref{tab:encoder}. It can be observed that the choice of backbones has a significant effect on results, and ViT-B/16 achieves the best performance, followed by RN50x4. Moreover, upgrading ResNet-50 to ResNet-101 with more parameters and deeper layers would only provide a little improvement for our task.

Furthermore, since we employ the pretrained ViT-B/16 as our visual encoder and use the feature map of all patches, we experiment with using the $cls$ token. Results show that PLCC and SRCC decrease 0.014 and 0.017 on LS-PCQA, respectively. By considering richer information, our approach tends to learn general quality features rather than only relying on the class token.

\begin{table}[H]
\begin{center}
  \begin{tabular}{ccccc}
    \toprule
    Backbone & PLCC$\uparrow$ & SRCC$\uparrow$ & RMSE$\downarrow$\\
    \midrule
    RN50 & 0.710 & 0.689 & 0.575 \\
    RN50x4 & 0.724 & 0.703 & 0.565 \\
    RN101 & 0.721 & 0.703 & 0.567 \\
    ViT-B/32 & 0.722 & 0.700 & 0.561 \\
    \textbf{ViT-B/16} & \textbf{0.755} & \textbf{0.736} & \textbf{0.533} \\
  \bottomrule
\end{tabular}
\caption{Performances of CLIP-PCQA for different backbones of visual encoder on LS-PCQA. RN50 denotes ResNet-50 and RN50x4 denotes ResNet-50 with 4 times more computations. ViT-B/32 represents vision transformer with 32 × 32 patch embeddings. The best performances are in \textbf{bold}.}
\label{tab:encoder}
\end{center}
\end{table}

\subsubsection{Number of projection views} 
During projection, we generate color map and their corresponding depth map for each viewpoint, which leads to the number of color maps matching the number of depth maps. To investigate the impact of viewpoint count, we test the model performance under different $M$ values. According to \Cref{tab:projection views}, we can see that employing 6 viewpoints yields the best performance. The reason may be that too few viewpoints hardly provide sufficient information to describe point clouds, while too many viewpoints may introduce redundant information.

\begin{table}[H]
\begin{center}
  \begin{tabular}{cccccc}
    \toprule
    M & 2 & 4 & \textbf{6} & 8 & 10 \\
    \midrule
    PLCC$\uparrow$ & 0.745 & 0.743 & \textbf{0.755} & 0.752 & 0.751 \\
    SRCC$\uparrow$ & 0.726 & 0.726 & \textbf{0.736} & 0.732 & 0.724 \\
    RMSE$\downarrow$ & 0.543 & 0.545 & \textbf{0.533} & 0.537 & 0.534 \\
  \bottomrule
\end{tabular}
\caption{Ablation of the number of viewpoints on LS-PCQA. The best performances are in \textbf{bold}.}
\label{tab:projection views}
\end{center}
\end{table}

\subsubsection{Prompt design} 
\begin{figure}[H]
\centering
\includegraphics[width=0.98\linewidth]{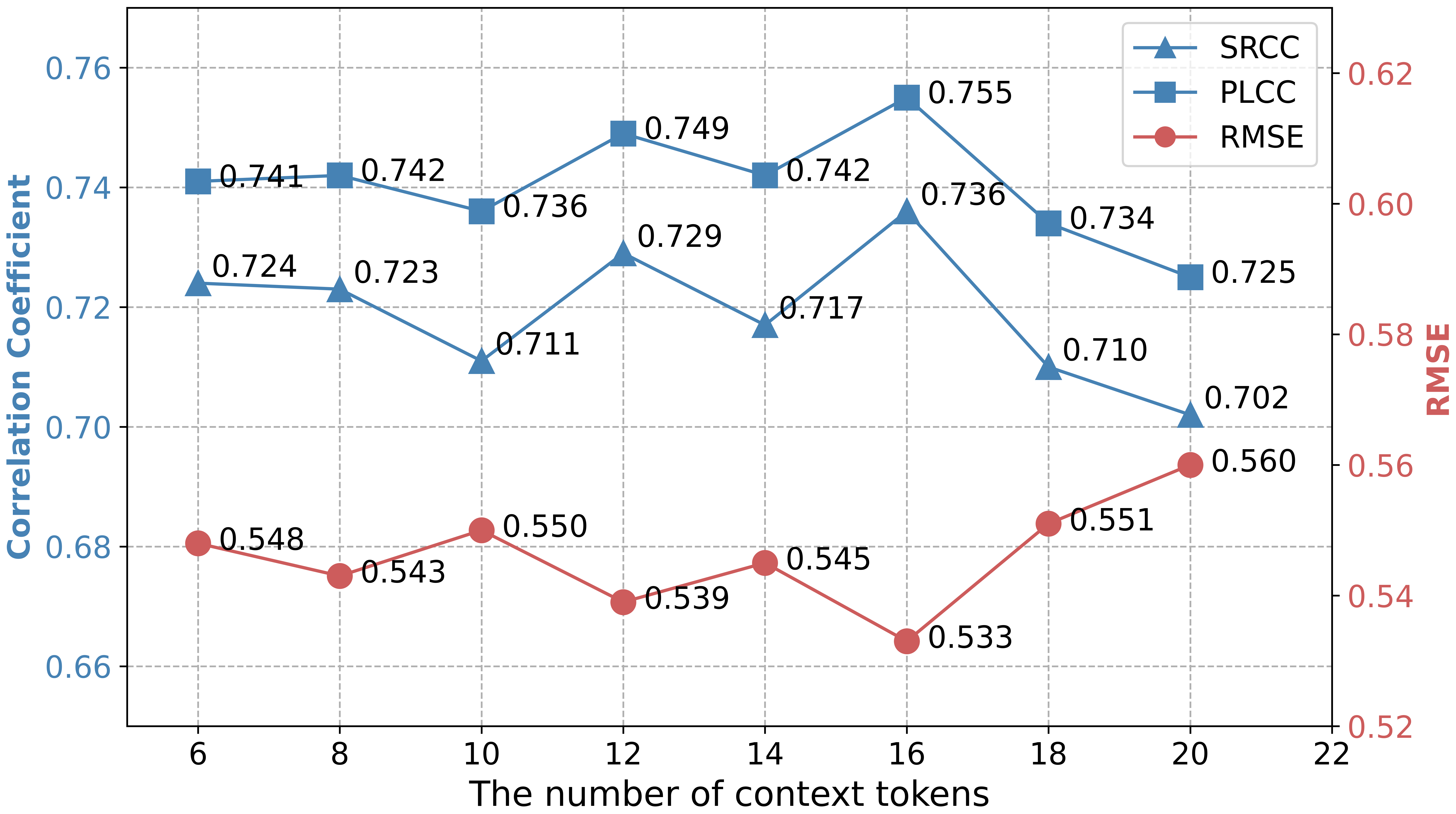}
\caption{The impact of the learnable prompt length on LS-PCQA. }
\label{fig:prompt_length_ablation}
\end{figure}

We explore the sensitivity of the prompt length $W$ in \Cref{fig:prompt_length_ablation}, where the model performs the best with prompt length of 16. Short prompts may constrain the learning space, potentially leading to the loss of crucial information. On the other hand, having much more context tokens might introduce unnecessary noise, which may obscure important information and degrade performance. Therefore, balancing the prompt length is critical to optimizing the model's performance.

\subsubsection{Distribution Prediction}
In this section, we aim to verify that predicting Opinion Score Distribution (OSD) is superior to directly predicting a single Mean Opinion Score 
(MOS). Specifically, during the training process, we calculate the final quality score by a linear combination of the predicted OSD and the score values of different quality levels, and we use Mean Squared Error (MSE) as loss function. The results, in terms of PLCC and SRCC, indicate a decrease of 0.011 and 0.005 on LS-PCQA, and a decrease of 0.011 and 0.028 on BASICS, respectively. These findings suggest that deriving the MOS from the predicted OSD yields more accurate results than predicting the MOS directly.

\begin{figure*}[t]
    \centering
    \begin{subfigure}[b]{0.98\textwidth}
        \centering
    \includegraphics[width=0.98\textwidth]{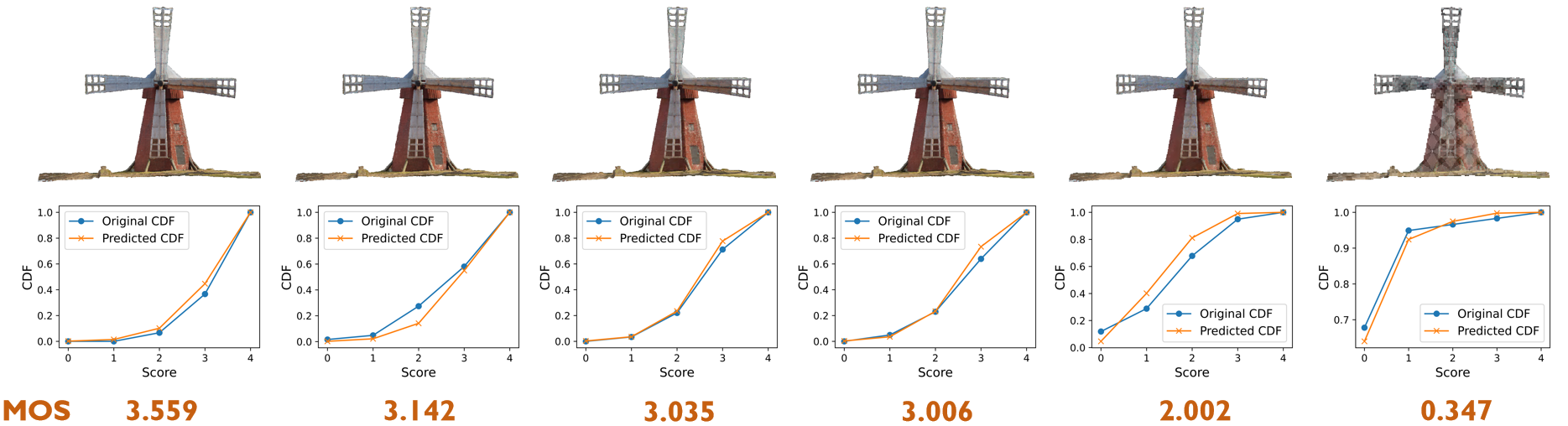}
    \caption{Comparison of the original Cumulative Distribution Function (CDF) and the CDF predicted by our method.}
    \label{fig:cdf}
    \end{subfigure}
    \begin{subfigure}[b]{0.98\textwidth}
        \centering
    \includegraphics[width=0.98\linewidth]{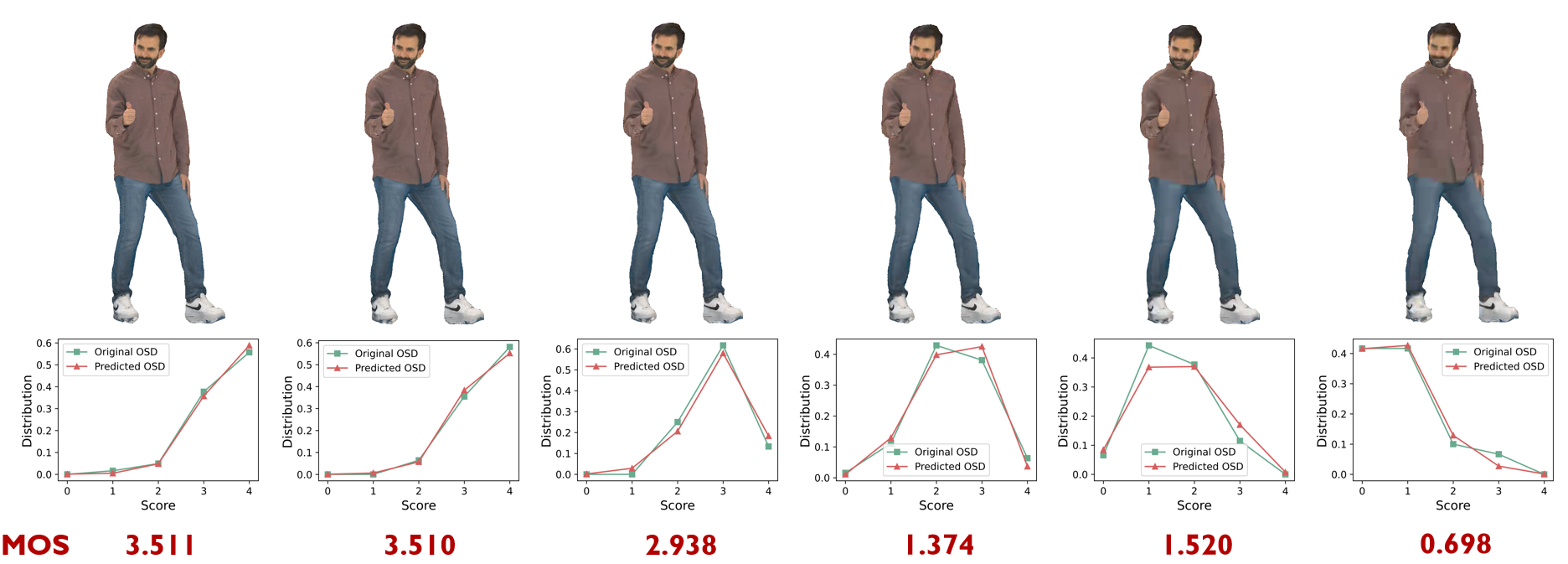}
    \caption{Comparison of the original Opinion Score Distribution (OSD) and the distribution predicted by our method.}
    \label{fig:osd}
    \end{subfigure}
\caption{Visualization of the original and predicted CDF or OSD.}
\label{fig:cdf_osd_compare}  
\end{figure*}
\subsection{Additional Visualization}
In this paper, we regard the OSD rather than a single Mean Opinion Score (MOS) as our final target. To demonstrate the effectiveness of our model in predicting the OSD, we provide a clear illustration in this section.

First, considering that our loss function includes the Earth Mover’s Distance (EMD) of the Cumulative Distribution Function (CDF), we compared the predicted CDF against the ground truth in \Cref{fig:cdf}. Additionally, since our objective is to predict the OSD in subjective experiments, we visualized the predicted OSD and the ground truth in \Cref{fig:osd}. The high consistency observed between the predicted and actual CDF or OSD demonstrates that, although our loss function does not directly constrain the OSD, our method effectively predicts it. This also validates that the similarities between visual and multiple textual features accurately capture the diversity and uncertainty inherent in subjective experiments.

\end{document}